\documentclass[12pt]{article}
\usepackage{acl2001}
\title{Correction of Errors in a Modality Corpus Used for 
Machine Translation by Using Machine-learning Method}
\author{Masaki Murata, Masao Utiyama, Kiyotaka Uchimoto, Qing Ma, \and Hitoshi Isahara \\
Communications Research Laboratory\\
2-2-2 Hikaridai, Seika-cho, Soraku-gun, Kyoto 619-0289, Japan\\
{\tt \{murata,mutiyama,uchimoto,qma,isahara\}@crl.go.jp}}

\begin{document}
\maketitle
\begin{abstract}
We performed 
corpus correction on a modality corpus for machine translation 
by using such machine-learning methods 
as the maximum-entropy method. 
We thus constructed a high-quality modality corpus 
based on corpus correction. 
We compared several kinds of methods 
for corpus correction in our experiments and 
developed a good method for corpus correction. 
\end{abstract}

%%%%%%%%%%%%%%%%%%%%%%%%%%%%%%%%%%%%%%
\section{Introduction}
%%%%%%%%%%%%%%%%%%%%%%%%%%%%%%%%%%%%%%

In recent years, 
various types of tagged corpora have been constructed 
and much research using tagged corpora has been performed. 
However, tagged corpora include errors, 
which impede the progress of research. 
Therefore, 
the correction of errors in corpora is an important research issue.\footnote{There is no previous paper 
on error correction in corpora. 
In terms of error detection in corpora, 
there has been research using boosting or anomaly detection \cite{Abney99,Eskin2000}.}

We have researched error correction in corpora 
by using the modality corpus we are currently constructing.\footnote{This paper is 
the English translation of the paper \cite{NLP2001_eng1}. 
We also performed corpus correction in a morphological corpus \cite{murata2000_3_nl_eng}.} 
This modality corpus consists of supervised learning data 
used for research on translating 
Japanese tense, aspect, and modality into English \cite{Murata_modal_tmi,murata_acl01_modal_appear}. 
(In this paper, we regard the word {\it modality} in 
the broad sense of including tense and aspect.) 
Tense, aspect, and modality are known to present difficult problems 
in machine translation. 
In traditional approaches, 
tense, aspect, and modality have been translated by using 
manually constructed heuristic rules. 
Recently, however, such corpus-based 
approaches as the example-based method have also been applied. 
The modality corpus we consider in this paper is necessary 
for such machine translation based on the corpus-based approach. 

In this paper, we describe the modality corpus 
in Section \ref{sec:corpus}, 
the method of corpus correction in Section \ref{sec:method}, 
and our experiments on corpus correction in Section \ref{sec:experiment}. 

\begin{figure}[t]
  \begin{center}
    \leavevmode
\begin{tabular}{|p{7cm}|}\hline
, {\it kono kodomo wa aa ieba kou iu kara koniku-rashii}\\
This child always talks back to me, and this \verb+<v>+is\verb+</v>+ why I \verb+<vj>+hate\verb+</vj>+ him.\\[0.2cm]
d {\it kare ga aa okubyou da to wa omowana-katta}\\
I \verb+<v>+did not think\verb+</v>+ he was so timid.\\[0.2cm]
c {\it aa isogashikute wa yasumu hima mo nai hazu da}\\
Such a busy man as he \verb+<v>+cannot have\verb+</v>+ any spare time.\\\hline
\end{tabular}
    \caption{Part of the modality corpus}
    \label{fig:modal_corpus}
\end{center}
\end{figure}

%%%%%%%%%%%%%%%%%%%%%%%%%%%%%%%%%%%%%%
\section{Modality Corpus for Machine Translation}
\label{sec:corpus}
%%%%%%%%%%%%%%%%%%%%%%%%%%%%%%%%%%%%%%

In this section, we describe the modality corpus. 
A part of it is shown in Figure \ref{fig:modal_corpus}. 
It is composed of a Japanese-English bilingual corpus, and 
each English sentence can include 
the following two types of tags. 
\begin{itemize}
\item 
The English main verb phrase is tagged with \verb+<v>+. 

\item 
The English verb phrase corresponding to 
the Japanese main verb phrase is tagged with \verb+<vj>+. 

\end{itemize}
The symbols at the beginning of 
each Japanese sentence, such as ``c'' and ``d'', indicate 
a category of tense, aspect, and modality for the sentence. 
(For example, ``c'' and ``d'' indicate
``can'' and ``past tense'', respectively. 
The first symbol in Figure \ref{fig:modal_corpus} 
is ``,''. 
This symbol is used when \verb+<vj>+ is used, such that 
the left part indicates the category of the verb phrase 
tagged with \verb+<v>+ and the right part indicates 
the category of the verb phrase tagged with \verb+<vj>+. 
In this corpus, the number of examples of present tense is large, so 
the symbol for present tense is a null expression (i.e., ``'').)
\verb+<vj>+ is tagged 
when the verb phrase with \verb+<v>+ 
does not correspond with the Japanese main verb. 

We use the following 34 categories for tense, aspect, and modality. 
These categories are determined by the surface expressions 
of the English verb phrases. 
{
\begin{enumerate}
\item 
  all combinations of   \{present tense, past tense\}, 
  \{progressive, not-progressive\}, and \{perfect, not-perfect\} (8 categories)
\item 
  imperative mood  (1 category)
\item 
  auxiliary verbs (\{present tense, past tense\} of ``be able to'', \{present tense, past tense\} of ``be going to'', ``can'', ``could'', \{present tense, past tense\} of ``have to'', ``had better'', ``may'', ``might'', ``must'', ``need'', ``ought'', ``shall'', ``should'', ``used to'', ``will'', ``would'') (19 categories)
\item 
  noun phrases (one category)
\item 
  participial construction (one category)
\item 
  verb ellipsis (one category)
\item 
  interjection or greeting sentences (one category)
\item 
  the case when a Japanese main verb phrase cannot correspond 
  to an English verb phrase (one category)
\item 
  the case when tagging cannot be performed (one category)
\end{enumerate}}
These categories of 
tense, aspect, and modality are defined on the basis of 
the surface expressions of the English sentences. 
So, if we can estimate the correct category from a Japanese sentence, 
we should be able to translate the Japanese tense, aspect, and modality 
into English. 
Therefore, in researching the translation of modality expressions based 
on the machine-learning method, 
only the tags indicating the categories of tense, aspect, and modality 
and the Japanese sentences are used. 

We placed an order with an outside company to construct 
the modality corpus according to the above conditions. 
We used about 40,000 example sentences 
from the Kodansha Japanese-English dictionary \cite{kodanshawaeiE} 
as a bilingual corpus. 
The outside company performed 
the tagging of \verb+<v>+ and 
the corresponding categories of modality by hand. 
Inspection work was performed more than twice, 
until the outside company considered 
no errors at all to exist in the corpus. 

%%%%%%%%%%%%%%%%%%%%%%%%%%%%%%%%%%%%%%
\section{Method of Corpus Correction}
\label{sec:method}
%%%%%%%%%%%%%%%%%%%%%%%%%%%%%%%%%%%%%%

In this section, 
we describe the method of correcting 
errors in the modality corpus constructed by hand, 
as described in the previous section. 
The method is to calculate the probabilities of tags, 
which are objects for error correction in a corpus, 
and then perform corpus correction by using 
those probabilities. 
In this paper, we only consider tags for modality categories, 
not ``\verb+<v>+'' and ``\verb+<vj>+'' tags. 

We tested two kinds of methods for calculating 
the probability of each tag: 
the maximum-entropy method, and the decision-list method.\footnote{In this paper, 
we use the maximum-entropy method and the decision-list method to 
calculate the probabilities of each tag. 
However, we may use a more accurate method 
to calculate the probabilities for corpus correction.}
{
\begin{itemize}
\item 
  Method based on the maximum-entropy method \cite{ristad97,ristad98}

  In this method, 
  the distribution of probabilities $p(a,b)$ 
  is calculated for the case 
  when Equation (\ref{eq:constraint}) is satisfied and 
  Equation (\ref{eq:entropy}) is maximized, 
  and the desired probabilities $p(a|b)$ 
  are then calculated by using the distribution of probabilities $p(a,b)$:

{\footnotesize
\begin{eqnarray}
  \label{eq:constraint}
  \sum_{a\in A,b\in B}p(a,b)g_{j}(a,b) 
  \ = \sum_{a\in A,b\in B}\tilde{p}(a,b)g_{j}(a,b)\\
  \ for\ \forall f_{j}\ (1\leq j \leq k) \nonumber
\end{eqnarray}
}
{\footnotesize
\begin{eqnarray}
  \label{eq:entropy}
  H(p) & = & -\sum_{a\in A,b\in B}p(a,b)\ log\left(p(a,b)\right),
\end{eqnarray}
}
where $A, B$, and $F$ are sets of categories, contexts, 
and features $f_j (\in F, 1\leq j\leq k)$, respectively; 
$g_{j}(a,b)$ is a function defined as 1 when context $b$ has feature $f_j$ 
and the category is $a$, or defined as 0 otherwise; 
and $\tilde{p}(a,b)$ is the occurrence rate of 
$(a,b)$ in the training data. 

In general, the distribution of $\tilde{p}(a,b)$ is very sparse. 
We cannot use it directly, 
so we must estimate the true distribution of $p(a,b)$ 
from the distribution of $\tilde{p}(a,b)$. 
We assume that the estimated values of the frequency of 
each pair of category and feature 
as calculated from $\tilde{p}(a,b)$ are the same as 
those from $p(a,b)$ (This corresponds to Equation (\ref{eq:constraint}).). 
These estimated values are not so sparse. 
We can thus use the above assumption for calculating $p(a,b)$. 
Furthermore, we maximize the entropy 
of the distribution of $\tilde{p}(a,b)$ to 
obtain one solution of $\tilde{p}(a,b)$, 
because only using Equation \ref{eq:constraint} produces 
many solutions for $\tilde{p}(a,b)$. 
Maximizing the entropy has the effect of making 
the distribution more uniform and 
is known to be a good solution for data sparseness problems. 

\item 
  Method based on the decision-list method \cite{Yarowsky:ACL94}

  In this method, 
  the probability of each category is calculated 
  by using one of features $f_j (\in F, 1\leq j\leq k)$. 
  The probability that 
  produces category $a$ in context $b$ is given 
  by the following equation: 
{
\begin{eqnarray}
  \label{eq:decision_list}
  p(a|b) = p(a|f_{max}),
\end{eqnarray}
}
such that $f_{max}$ is defined by 
{
\begin{eqnarray}
  \label{eq:decision_list2}
  f_{max} = argmax_{f_j\in F} \ max_{a_i\in A} \ \tilde{p}(a_i|f_j),
\end{eqnarray}
}
where $\tilde{p}(a_i|f_j)$ is the occurrence rate of 
category $a_i$ when the context has feature $f_j$. 

%%Although the decision-list method is simple, 
%%it estimates probabilities on the basis of only one feature, 
%%so it performs poorly as a machine-learning method.  
\end{itemize}}

In this paper, we used the following items as 
features, which are the context 
when the probabilities are calculated. 
(26 ($= 5 + 10 + 10 + 1$) features appear 
in each English sentence.)
\begin{itemize}
\item 
  The strings of 1-gram to 5-gram just to the left of 
  \verb+<v>+ in the sentence. 

  (e.g.) \underline{I} \verb+<v>+did not think\verb+</v>+ he was so timid.
\item 
  The strings of 1-gram to 10-gram just to the right of \verb+<v>+. 

  (e.g.) I \verb+<v>+\underline{did not} think\verb+</v>+ he was so timid.
\item 
  The strings of 1-gram to 10-gram just to the left of \verb+</v>+. 

  (e.g.) I \verb+<v>+did \underline{not think}\verb+</v>+ he was so timid.
\item 
  The 1-gram string at the end of 
  the sentence. 

  (e.g.) I \verb+<v>+did not think \verb+</v>+ he was so timid\underline{.}
\end{itemize}
When the verb phrase was divided into two parts, 
as in an interrogative sentence, 
the above extraction of features was performed 
after eliminating the words between the first \verb+</v>+ and the second \verb+<v>+. 

Because the corpus used in this paper was designed for estimating 
the modality of the English sentence from the Japanese sentence, 
one may think that we should extract the features from the Japanese sentence. 
It is true if we want to infer English modalities from Japanese sentences. 
What we want to do is, however, to correct 
English modality tags. Thus we should use all the information available. 
Since the category of the modality expression 
of the English sentence is tagged and 
the verb phrase of the English sentence is examined 
for construction of the corpus by hand, 
it is reasonable to use 
the English verb phrase 
in corpus correction based on the machine-learning method. 

Next, we describe the method of judging 
whether each tag in the corpus is incorrect or not. 
We first calculate the probabilities of 
the category of the tag, and of the other categories. 
We judge that the tag is correct 
when its category has the 
highest probability and 
incorrect 
when one of the other categories has the 
highest probability. 
Next, we correct the tag 
if it is judged to be incorrect. 
This correction is performed by changing 
the tag to the tag of the category 
with the highest probability. 
(This correction is confirmed by annotators in actuality.) 
\footnote{This action of corpus correction is 
exactly equivalent to 
redefining the tag in the corpus 
by using a machine-learning method and 
re-tagging the newly defined tag.}

Corpus correction should be confirmed by human beings. 
Therefore it is very time consuming. 
However, when the probabilities of each tag can be calculated, 
we can define the confidence value of the corpus correction, 
as described below. 
It is thus more convenient 
to sort the error candidates in the corpus 
by confidence value and 
begin by correcting the errors for which 
the confidence value is higher. 

We tested the following two types of methods 
for determining the confidence value for corpus correction. 
{
\begin{itemize}
\item 
  {\bf Method 1} --- the probability of the category 
  with the highest probability is used as 
  the confidence value for corpus correction. 

\item 
  {\bf Method 2} --- the non-probability of the tag 
  originally defined is used as 
  the confidence value for corpus correction. 

\end{itemize}}
In this paper, 
the non-probability is defined as the value obtained 
by subtracting the probability from 1. 
%%%Method 1 is when 
%%%we consider that 
%%%the case where the tag given to the corpus after the corpus correction 
%%%has a higher probability is a better corpus correction and 
%%%Method 2 is when 
%%%we consider that 
%%%the case where the tag originally given to the corpus before the corpus correction 
%%%has a lower probability is a better corpus correction. 

%%We have explained almost all the methods used for corpus correction. 
We finally explain the methods to use 
data for calculating probabilities. 
There are two kinds of methods for 
calculating the probabilities 
by using the machine-learning method: 
{
\begin{itemize}
\item 
  calculation of probabilities for the closed data, and  

\item 
  calculation of probabilities for the open data.

\end{itemize}}
The first method calculates probabilities by using 
all the tags in the corpora 
including the tag which is judged currently. 
The second method does not use 
the tag which is judged currently. 
In this paper, 10-fold cross validation was used for 
calculating probabilities for the open data. 
\footnote{When the probabilities are calculated using 
open data in the decision-list method, 
the probability of the category of the original tag is 
apt to be 0, or the probability of the category of the tag 
defined after corpus correction is apt to be 1, because 
the calculation is performed by not using 
the original tag. Thus when there are many such tags, 
many of them have the same probability and 
sorting by probabilities becomes difficult. 
In this case, we sort the tags 
by arranging those whose probability is calculated 
from the features which have many tags 
in descending order of confidence value for corpus correction.}

%%%%%%%%%%%%%%%%%%%%%%%%%%%%%%%%%%%%%%
\section{Experiments on Corpus Correction}
\label{sec:experiment}
%%%%%%%%%%%%%%%%%%%%%%%%%%%%%%%%%%%%%%

We carried out experiments on corpus correction 
by using the methods described in the previous section. 
These experiments were performed after eliminating 
the sentences given tags 
indicating that tagging could not be performed. 
Thus, 
these experiments were performed for 39,718 modality tags. 
The results are shown in Tables \ref{tab:me_result_close} to \ref{tab:ds_result_open}. 
``random 300'' indicates 
the precisions for 300 tags extracted randomly from among the tags 
corrected by our system. 
``top X'' indicates 
the precisions for the top X tags sorted 
by Method 1 or Method 2. 
``Precision for detection'' indicates 
the percentage of tags for which 
detection of an error succeeded in causing the tag 
to be corrected by our system, while 
``Precision for correction'' indicates 
the percentage of tags for which 
correction of an error succeeded in causing the tag 
to be corrected by our system. 

We came to the following conclusions based on the experimental results.
{
\begin{itemize}
\item 
  Throughout all the experiments, 
  the precisions for detection and 
  correction were almost the same. 
  Thus, we found that 
  it is more convenient to perform both 
  correction and detection, 
  rather than only detection. 

  From the viewpoint of manual modification, 
  when we modify tags by hand, 
  it is also more convenient for 
  the system to produce a candidate category 
  that is tagged to the corpus after corpus correction. 
  This is because we can find how the original tag 
  was incorrect and how we should change it to 
  the new corrected tag. 
  When only detection is performed, in other words, 
  a candidate category is not presented, 
  an annotator may not know 
  why the tag is incorrect. 

\item 
  In general, the maximum-entropy method produced 
  higher precision than the decision-list method. 
  However, when the closed data was used to 
  the calculate the probabilities, 
  the precisions of the top items 
  were almost the same for 
  the two methods. 

\item 
  In terms of the precisions of top items, 
  using the closed data to calculate 
  the probabilities was better than 
  using the open data. 
  However, 
  in terms of the total number of extracted items, 
  using the open data was better. 

\item 
  In terms of sorting by Method 1 or Method 2, 
  Method 1 generally produced higher precisions for the top items 
  than Method 2. 

\item 
  In terms of comparing ``random 300'' and ``top X'', 
  ``top X'' produced much higher precisions for the top items 
  than ``random 300''.   We thus found that 
  sorting by confidence values of corpus correction 
  is very important. 

\end{itemize}}

\begin{table*}[p]
\vspace*{-.5cm}
  \caption{Precision of corpus correction using the maximum-entropy method 
    (The probabilities were calculated using the closed data. 184 candidate errors were extracted.)}
    \label{tab:me_result_close}
  \begin{center}
\begin{tabular}[c]{|c|lr||rc|rc|}\hline
\multicolumn{3}{|c||}{}  & \multicolumn{2}{|c|}{Precision for detection} & \multicolumn{2}{|c|}{Precision for correction} \\\hline
\multicolumn{3}{|c||}{random 300} & 69\% &(127/184) & 68\% & (126/184) \\\hline
Method 1 &top&  50 & 100\% & ( 50/ 50) & 100\% & ( 50/ 50) \\
 &top& 100 &  92\% & ( 92/100) &  92\% & ( 92/100) \\
 &top& 150 &  77\% & (116/150) &  77\% & (116/150) \\
   &top& 200 &  69\% & (127/184) &  68\% & (126/184) \\
   &top& 250 &  ---  & ---       &  ---  & ---       \\
   &top& 300 &  ---  & ---       &  ---  & ---       \\\hline
Method 2 &top&  50 &  88\% & ( 44/ 50) &  88\% & ( 44/ 50) \\ 
 &top& 100 &  81\% & ( 81/100) &  81\% & ( 81/100) \\ 
 &top& 150 &  74\% & (112/150) &  74\% & (111/150) \\ 
   &top& 200 &  69\% & (127/184) &  68\% & (126/184) \\ 
   &top& 250 &  ---  & ---       &  ---  & ---       \\ 
   &top& 300 &  ---  & ---       &  ---  & ---       \\\hline
\end{tabular}
\end{center}
\end{table*}

\begin{table*}[p]
\vspace*{-.5cm}
  \caption{Precision of corpus correction using the maximum-entropy method (The probabilities were calculated using the open data. 694 candidate errors were extracted.)}
    \label{tab:me_result_open}
  \begin{center}
\begin{tabular}[c]{|c|lr||rc|rc|}\hline
\multicolumn{3}{|c||}{}  & \multicolumn{2}{|c|}{Precision for detection} & \multicolumn{2}{|c|}{Precision for correction} \\\hline
\multicolumn{3}{|c||}{random 300} &  28\% &( 84/300) & 26\% &( 78/300) \\\hline
Method 1 &top&  50 &  88\% & ( 44/ 50) &  88\% & ( 44/ 50) \\
 &top& 100 &  88\% & ( 88/100) &  88\% & ( 88/100) \\
 &top& 150 &  80\% & (121/150) &  79\% & (119/150) \\
   &top& 200 &  68\% & (136/200) &  67\% & (134/200) \\
   &top& 250 &  60\% & (151/250) &  59\% & (149/250) \\
   &top& 300 &  53\% & (160/300) &  52\% & (157/300) \\\hline
Method 2 &top&  50 &  72\% & ( 36/ 50) &  72\% & ( 36/ 50) \\ 
 &top& 100 &  74\% & ( 74/100) &  71\% & ( 71/100) \\ 
 &top& 150 &  70\% & (106/150) &  68\% & (102/150) \\ 
   &top& 200 &  67\% & (135/200) &  65\% & (131/200) \\ 
   &top& 250 &  60\% & (152/250) &  58\% & (147/250) \\ 
   &top& 300 &  52\% & (157/300) &  50\% & (152/300) \\\hline
\end{tabular}
\end{center}
\end{table*}

\begin{table*}[p]
\vspace*{-.5cm}
  \caption{Precision of corpus correction using the decision-list method 
    (The probabilities were calculated using the closed data. 383 candidate errors were extracted.)}
    \label{tab:ds_result_close}
  \begin{center}
\begin{tabular}[c]{|c|lr||rc|rc|}\hline
\multicolumn{3}{|c||}{}  & \multicolumn{2}{|c|}{Precision for detection} & \multicolumn{2}{|c|}{Precision for correction} \\\hline
\multicolumn{3}{|c||}{random 300} & 34\% & (104/300) &  33\% & (101/300) \\\hline
Method 1  &top&  50 & 100\% & ( 50/ 50) & 100\% & ( 50/ 50) \\      
 &top& 100 &  92\% & ( 92/100) &  92\% & ( 92/100) \\      
 &top& 150 &  76\% & (115/150) &  74\% & (112/150) \\      
   &top& 200 &  62\% & (124/200) &  60\% & (121/200) \\      
   &top& 250 &  51\% & (128/250) &  50\% & (125/250) \\      
   &top& 300 &  44\% & (132/300) &  43\% & (129/300) \\\hline
Method 2 &top&  50 &  88\% & ( 44/ 50) &  86\% & ( 43/ 50) \\      
 &top& 100 &  86\% & ( 86/100) &  84\% & ( 84/100) \\      
 &top& 150 &  71\% & (107/150) &  69\% & (104/150) \\      
   &top& 200 &  59\% & (118/200) &  57\% & (115/200) \\      
   &top& 250 &  50\% & (126/250) &  49\% & (123/250) \\      
   &top& 300 &  43\% & (129/300) &  42\% & (126/300) \\\hline
\end{tabular}
\end{center}
\end{table*}

\begin{table*}[p]
\vspace*{-.5cm}
  \caption{Precision of corpus correction using the decision-list method (the probabilities were calculated using the open data. 694 candidate errors were extracted.)}
    \label{tab:ds_result_open}
  \begin{center}
\begin{tabular}[c]{|c|lr||rc|rc|}\hline
\multicolumn{3}{|c||}{}  & \multicolumn{2}{|c|}{Precision for detection} & \multicolumn{2}{|c|}{Precision for correction} \\\hline
\multicolumn{3}{|c||}{random 300} &  6\% & ( 18/300) & 6\% & ( 18/300) \\\hline
Method 1 &top&  50 &  56\% & ( 28/ 50) & 52\% & ( 26/ 50) \\      
 &top& 100 &  43\% & ( 43/100) & 40\% & ( 40/100) \\      
 &top& 150 &  31\% & ( 47/150) & 29\% & ( 44/150) \\      
   &top& 200 &  26\% & ( 52/200) & 24\% & ( 48/200) \\      
   &top& 250 &  22\% & ( 55/250) & 20\% & ( 51/250) \\      
   &top& 300 &  20\% & ( 61/300) & 19\% & ( 57/300) \\\hline
Method 2 &top&  50 &  66\% & ( 33/ 50) & 64\% & ( 32/ 50) \\      
 &top& 100 &  48\% & ( 48/100) & 46\% & ( 46/100) \\      
 &top& 150 &  44\% & ( 66/150) & 42\% & ( 63/150) \\      
   &top& 200 &  35\% & ( 71/200) & 34\% & ( 68/200) \\      
   &top& 250 &  30\% & ( 77/250) & 29\% & ( 73/250) \\      
   &top& 300 &  26\% & ( 80/300) & 25\% & ( 76/300) \\\hline
\end{tabular}
\end{center}
\end{table*}

Based on the above results, 
we think that the following strategy is a better solution. 

  \begin{enumerate}
  \item 
    We first perform high-quality corpus correction 
    by using the probability calculation for the closed data 
    and Method 1. 

  \item 
    Next, we perform corpus correction for 
    a much larger number of tags 
    by using the probability calculation for the open data, 
    the maximum-entropy method, and Method 1. 
  \end{enumerate}

%%%%%%%%%%%%%%%%%%%%%%%%%%%%%%%%%%%%%%
\section{Conclusion}
%%%%%%%%%%%%%%%%%%%%%%%%%%%%%%%%%%%%%%

In this paper, 
we have described 
corpus correction using a machine-learning method 
for a modality corpus for machine translation. 
We have constructed a high-quality modality corpus 
by using corpus correction. 
In the future, we plan 
to research Japanese-English translation 
of tense, aspect, and modality by using 
this corpus. 

Our method of corpus correction has 
the following advantages. 
{
\begin{itemize}
\item 
  There is no previous paper
  on error correction in corpora. 

  In terms error detection in corpora, 
  there has been research using boosting or anomaly detection \cite{Abney99,Eskin2000}. 
  We found that 
  the precisions for detection and 
  correction were almost the same. 
  Therefore, we should 
  perform correction in addition to detection. 

\item 
  Our method calculates the probability of each tag
  and can sort the error candidates in the corpus 
  by using these probabilities as confidence values for corpus correction. 
  Thus, 
  we can begin to correct errors for which 
  the confidence value is higher. 

\item
  Our method uses the machine-learning method and 
  inherits its original advantages. 
  \begin{itemize}
  \item 
    Our method has the same wide applicability as 
    the machine-leaning method and 
    can be used to correct a various types of corpora. 

  \item 
    A large amount of human effort is not necessary, and 
    human beings only have to provide appropriate feature sets used in the machine-learning method. 

  \end{itemize}
\end{itemize}}

%%%%%%%%%%%%%%%%%%%%
%%% Bibliography %%%
%%%%%%%%%%%%%%%%%%%%
{

}
\end{document}